\RequirePackage{fix-cm}
\documentclass[smallextended]{svjour3}\usepackage[]{graphicx}\usepackage[]{color}
\makeatletter
\def\maxwidth{ %
  \ifdim\Gin@nat@width>\linewidth
    \linewidth
  \else
    \Gin@nat@width
  \fi
}
\makeatother

\definecolor{fgcolor}{rgb}{0.345, 0.345, 0.345}

\usepackage{framed}
\makeatletter
\newenvironment{kframe}{%
 \def\at@end@of@kframe{}%
 \ifinner\ifhmode%
  \def\at@end@of@kframe{\end{minipage}}%
  \begin{minipage}{\columnwidth}%
 \fi\fi%
 \def\FrameCommand##1{\hskip\@totalleftmargin \hskip-\fboxsep
 \colorbox{shadecolor}{##1}\hskip-\fboxsep
     \hskip-\linewidth \hskip-\@totalleftmargin \hskip\columnwidth}%
 \MakeFramed {\advance\hsize-\width
   \@totalleftmargin\z@ \linewidth\hsize
   \@setminipage}}%
 {\par\unskip\endMakeFramed%
 \at@end@of@kframe}
\makeatother

\definecolor{shadecolor}{rgb}{.97, .97, .97}
\definecolor{messagecolor}{rgb}{0, 0, 0}
\definecolor{warningcolor}{rgb}{1, 0, 1}
\definecolor{errorcolor}{rgb}{1, 0, 0}
\newenvironment{knitrout}{}{} 

\usepackage{alltt}       
\smartqed  
\usepackage{graphicx}
\usepackage{amsmath,amssymb,array}
\usepackage{booktabs}
\usepackage{todonotes}
\usepackage[sectionbib,round]{natbib}
\usepackage[hyphens]{url}
\usepackage{hyperref}

\newcommand{\new}[1]{{#1}}
\newcommand{\code}[1]{\texttt{#1}}

\newcounter{mycomment}

\IfFileExists{upquote.sty}{\usepackage{upquote}}{}
\begin{document}


\title{\new{\code{OpenML}: An R Package to Connect to the Machine Learning Platform OpenML}}
\titlerunning{An R Package to Connect to the OpenML Platform}



\author{Giuseppe~Casalicchio~\and~Jakob~Bossek~\and~Michel~Lang~\and~
Dominik~Kirchhoff~\and~Pascal~Kerschke~\and~Benjamin~Hofner~\and  
Heidi~Seibold~\and~Joaquin~Vanschoren~\and~Bernd~Bischl}


\institute{Giuseppe Casalicchio, Bernd Bischl \at
  Department of Statistics, Ludwig-Maximilians-University Munich, \\
  80539 Munich, Germany \\
  \email{giuseppe.casalicchio@stat.uni-muenchen.de}
\and
  Jakob Bossek, Pascal Kerschke \at
  Information Systems and Statistics, University of M{\"u}nster, \\
  48149 M{\"u}nster, Germany
\and
  Michel Lang \at
  Department of Statistics, TU Dortmund University, 
  44227 Dortmund, Germany
\and
  Dominik Kirchhoff \at
  Dortmund University of Applied Sciences and Arts, 
  44227 Dortmund, Germany
\and
  Benjamin Hofner \at
  Section of Biostatistics, Paul-Ehrlich-Institut, 
  63225 Langen, Germany
\and
  Heidi Seibold \at
  Epidemiology, Biostatistics and Prevention Institute, University of Zurich, \\
  8001 Zurich, Switzerland
\and
  Joaquin Vanschoren \at
  Eindhoven University of Technology, 
  5600 MB Eindhoven, Netherlands
}

\date{Received: date / Accepted: date}
\maketitle

\abstract{
OpenML is an online machine learning platform where researchers can
easily share data, machine learning tasks and experiments as well as organize them
online to work and collaborate more efficiently. In this paper, we present an
\code{R} package to interface with the OpenML platform and illustrate its usage in
combination with the machine learning \code{R} package \code{mlr}~\citep{JMLR:v17:15-066}.
We show how the \code{OpenML} package allows \code{R} users to easily search,
download and upload data sets and machine learning tasks. Furthermore, we also
show how to upload results of experiments, share them with others and
download results from other users. 
Beyond ensuring
reproducibility of results, the OpenML platform automates much of the drudge work, speeds up
research, facilitates collaboration and increases the users' visibility online.
}

\section{Introduction}
\label{sec:introduction}

OpenML is an online machine learning platform for sharing and organizing data,
machine learning algorithms and experiments \citep{openml2013}. It is designed to
create a frictionless, networked ecosystem~\citep{nielsen2012reinventing}, allowing people all over the world to collaborate and build directly on each
other's latest ideas, data and results. Key elements of OpenML are data sets,
tasks, flows and runs:

\begin{itemize}
\item \textbf{Data sets} can be shared (under a licence) by uploading them or simply linking to existing data repositories (e.g., \href{http://mldata.org}{mldata.org}, \href{https://figshare.com}{figshare.com}).
For known data formats (e.g., ARFF for tabular data), OpenML will automatically
analyze and annotate the data sets with measurable characteristics to support detailed search and further analysis.
Data sets can be repeatedly updated or changed and are then automatically versioned.

\item \textbf{Tasks} can be viewed as containers including a data set and additional information defining what is to be learned. They define which input data
are given and which output data should be obtained. For instance, classification
tasks will provide the target feature, the evaluation measure (e.g., the area under the curve) and the estimation procedure (e.g., cross-validation splits) as inputs. As output they expect a
description of the machine learning algorithm or workflow that was used and, if available, its predictions.

\item \textbf{Flows} are implementations of single machine learning algorithms
or whole workflows that solve a specific task, e.g., a random forest 
implementation is a flow that can be used
to solve a classification or regression task. 
Ideally, flows are already implemented (or custom) algorithms in existing software that 
take OpenML tasks as inputs and can automatically read and solve them.
They also contain a list (and description) of possible hyperparameters that are available for the algorithm.

\item \textbf{Runs} are the result of executing flows, optionally with preset
hyperparameter values, on tasks and contain all expected outputs and evaluations
of these outputs (e.g., the accuracy of predictions).
Runs are fully reproducible because they are automatically linked to specific 
data sets, tasks, flows and hyperparameter settings.
They also include the authors of the run and any additional information provided 
by them, such as runtimes. 
Similar to data mining challenge platforms~\citep[e.g., Kaggle;][]{Carpenter:2011p34283}, OpenML evaluates all submitted results (using a range of evaluation measures) and
compares them online. The difference, however, is that
OpenML is designed for collaboration rather than competition: anyone can browse,
immediately build on and extend all shared results.
\end{itemize}

\noindent As an open science platform, OpenML provides important benefits for the science community and beyond.

\paragraph{Benefits for Science:}
Many sciences have made significant breakthroughs by adopting online tools that
help organizing, structuring and analyzing scientific data online~\citep{nielsen2012reinventing}.
Indeed, any shared idea, question, observation
or tool may be noticed by someone who has just the right expertise to
spark new ideas, answer open questions, reinterpret observations or reuse data
and tools in unexpected new ways. Therefore, sharing research results and 
collaborating online as a (possibly cross-disciplinary) team enables scientists
to quickly build on and extend the results of others, fostering new discoveries. 

Moreover, ever larger studies
become feasible as a lot of data are already available. 
Questions such as ``Which hyperparameter is important to tune?'', ``Which
is the best known workflow for analyzing this data set?'' or ``Which data sets are
similar in structure to my own?'' can be answered in minutes by
reusing prior experiments, instead of spending days setting up and running new
experiments~\citep{Vanschoren12}.


\paragraph{Benefits for Scientists:}
Scientists can also benefit personally from using Open\-ML. For example, they can 
\textit{save time}, because OpenML assists in many routine and tedious duties:
finding data sets, tasks, flows and prior results, setting up experiments
and organizing all experiments for further analysis. Moreover, new experiments
are immediately compared to the state of the art without always having to rerun
other people's experiments.

Another benefit is that linking one's results to those of others has a large potential for
\textit{new discoveries}~\new{\citep{feurer2015initializing, post2016does, Au2016}}, leading to more publications and more collaboration with
other scientists all over the world.
Finally, OpenML can help scientists to \textit{reinforce their reputation} by making their
work (published or not) visible to a wide group of people and by showing
how often one's data, code and experiments are downloaded or reused in the
experiments of others.

\paragraph{Benefits for Society:}
OpenML also provides a useful learning and working environment for students, citizen
scientists and practitioners. Students and citizen scientist can easily
explore the state of the art and work together with top minds by contributing
their own algorithms and experiments. Teachers can challenge their students by
letting them compete on OpenML tasks or by reusing OpenML data in
assignments. Finally, machine learning practitioners can explore and reuse the
best solutions for specific analysis problems, interact with the scientific
community or efficiently try out many possible approaches. \\

\noindent The remainder of this paper is structured as follows. First, we discuss the web
services offered by the OpenML server and the website on \href{http://openml.org}{OpenML.org} that allows
web access to all shared data and several tools for data organization and
sharing. Second, we briefly introduce the \code{mlr} package \citep{JMLR:v17:15-066, schiffner2016mlr}, which is a machine 
learning toolbox for \code{R} \citep{r2016} and offers a unified interface to many machine learning algorithms.
Third, we discuss and illustrate some important functions of the \code{OpenML R}
package. After that, we illustrate its usage in combination with
the \code{mlr R} package by conducting a short case study. Finally, we conclude
with a discussion and an outlook to future developments.

\section{The OpenML Platform}
\label{sec:server}

The OpenML platform consists of several layers of software:

\paragraph{Web API:}
Any application (or web application), can communicate with the OpenML server 
through the extensive Web API, 
\new{
an application programming 
interface (API) that offers a set of calls (e.g., to download/upload data) 
using representational state transfer (REST) which is a simple, lightweight 
communication mechanism based on standard HTTP requests. 
}
Data sets, tasks, 
flows and runs can be created, read, updated, deleted, searched and tagged 
through simple HTTP calls. An overview of calls is available on 
\url{http://www.openml.org/api_docs}.

\begin{figure}[p]
\begin{center}
\noindent \includegraphics[width=.98\linewidth]{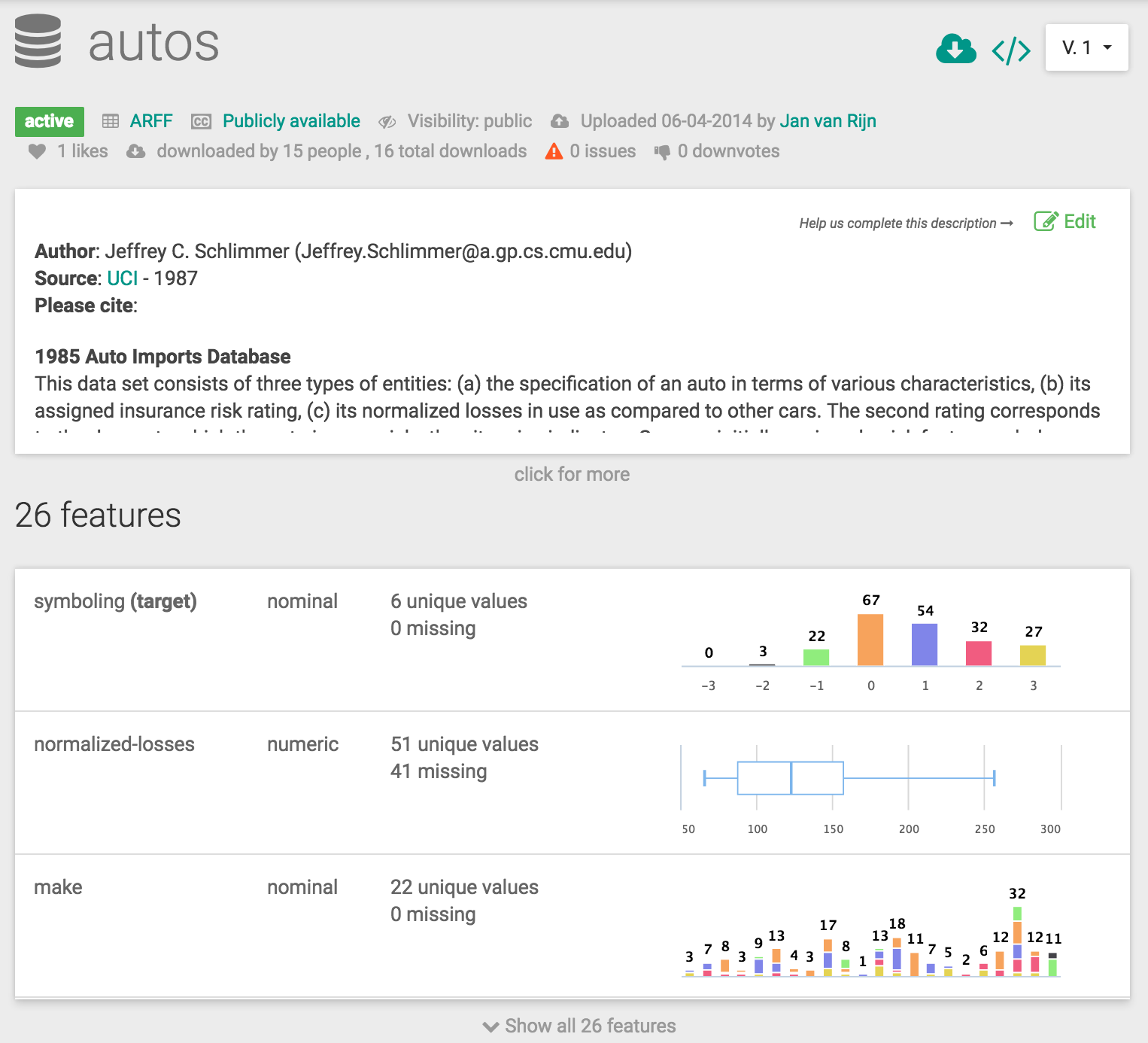}
\includegraphics[width=.98\linewidth]{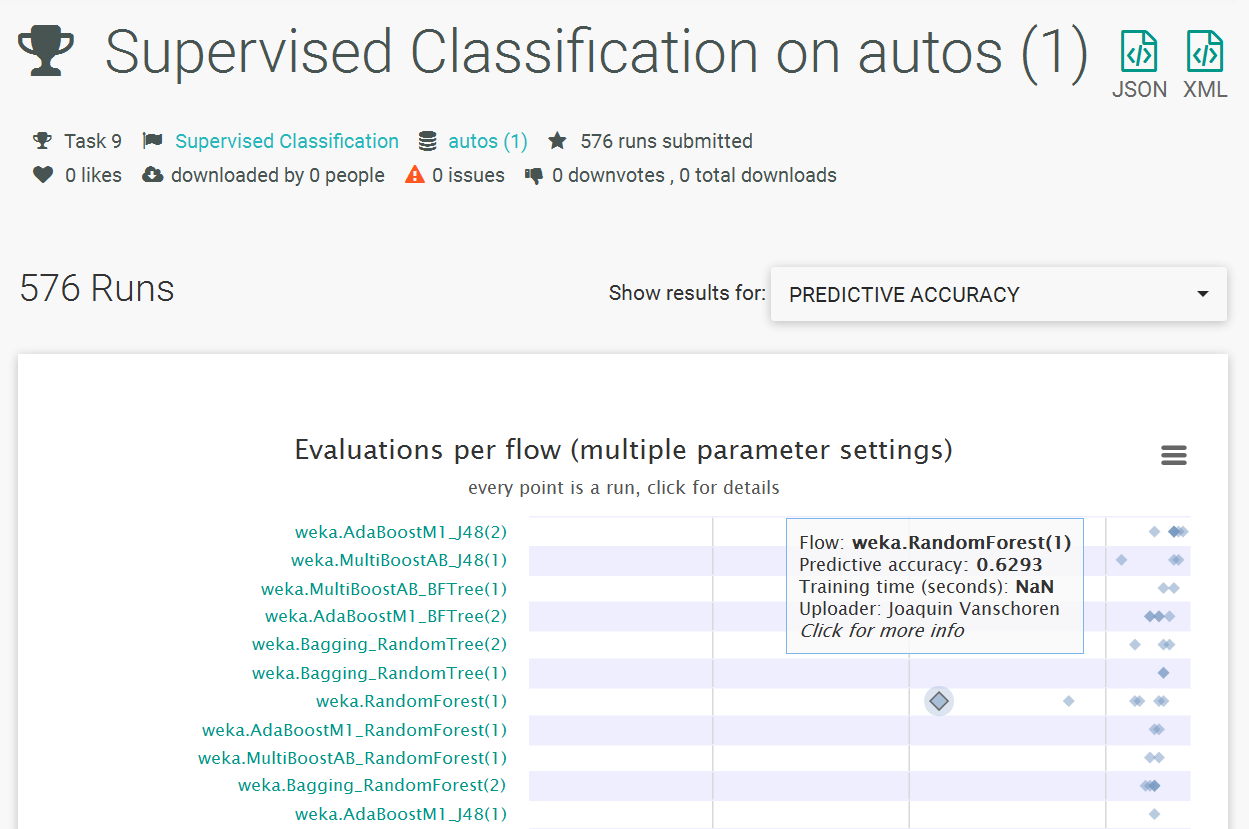}
\caption{Screenshots of the OpenML website. The top part shows the data set 
'autos', with wiki description and descriptive overview of the data features. 
The bottom part shows a classification task, with an overview of the best 
submitted flows with respect to the predictive accuracy as performance measure. 
Every dot here is a single run (further to the right is better).}
\label{screenshot}
\end{center}
\end{figure}

\paragraph{Website:}
\href{http://openml.org}{OpenML.org} is a website offering easy browsing, 
organization and sharing of all data, code and experiments. It allows users to 
easily search and browse all shared data sets, tasks, flows and runs, as well 
as to compare and visualize all combined results. It provides an easy way to 
check and manage your experiments anywhere, anytime and discuss them with 
others online. See Figure \ref{screenshot} for a few screenshots of the OpenML 
website.

\paragraph{
\new{
Programming Interfaces:
}
}
OpenML also offers interfaces in multiple programming languages, such as the 
\code{R} interface presented here, which hides the API calls and allow 
scientists to interact with the server using language-specific functions. As we 
demonstrate below, the \code{OpenML R} package allows \code{R} users to search 
and download data sets and upload the results of machine learning experiments in 
just a few lines of code. Other interfaces exist for \code{Python}, \code{Java} 
and \code{C\# (.NET)}. For tools that usually operate through a graphical 
interface, such as  WEKA~\citep{Hall:2009p14495}, MOA~\citep{Bifet:2010p28524} 
and RapidMiner~\citep{RCOMM2013}, plug-ins exist that integrate OpenML sharing 
facilities.\\

\noindent OpenML is organized as an open source project, hosted on 
GitHub 
\new{
(\url{https://github.com/openml}) 
}
and is free to use under the 
CC-BY licence. When uploading new data sets and code, users can select under 
which licence they wish to share the data, OpenML will then state licences and 
citation requests online and in descriptions downloaded from the Web API. 

OpenML has an active developer community and everyone is welcome to 
help extend it or post new suggestions through the website or through GitHub.
Currently, there are close to $1\,700\,000$ runs on about $20\,000$ data 
sets and $3\,500$ unique flows available on the OpenML platform. While still 
in beta development, it has over 
$1\,400$~registered users, over $1\,800$ frequent visitors and the website is 
visited by around 200 unique visitors every day, from all over the world. It 
currently has server-side support for classification, regression, clustering, 
data stream classification, learning curve analysis, survival analysis and 
machine learning challenges for classroom use.

\section{The mlr R Package}
\label{sec:mlr}


The \code{mlr} package \citep{JMLR:v17:15-066, schiffner2016mlr} offers a clean, easy-to-use and flexible
domain-specific language for machine learning experiments in \code{R}. An
object-\new{oriented} interface is adopted to unify the definition of machine
learning tasks, setup of learning algorithms, training of models, predicting
and evaluating the algorithm's performance. This unified interface hides the
actual implementations of the underlying learning algorithms. Replacing one
learning algorithm with another becomes as easy as changing a string. Currently,
\code{mlr} has built-in support for classification,
regression, multilabel classification, clustering and survival analysis and includes in total 160 modelling techniques.
\new{
A complete list of the integrated learners and how to integrate own learners, as well as
further information on the \code{mlr} package can be found in the corresponding
tutorial (\url{http://mlr-org.github.io/mlr-tutorial/}).
}
A plethora of further functionality is implemented in \code{mlr}, e.g., assessment of
generalization performance, comparison of different algorithms in a
scientifically rigorous way, feature selection and algorithms for hyperparameter tuning,
\new{including Iterated F-Racing \citep{lang2015automatic} and Bayesian optimization with the package
  \code{mlrMBO}~\citep{mlrMBO}}.
On top of that, \code{mlr} offers a wrapper mechanism, which
allows to extend learners through pre-train, post-train, pre-predict and
post-predict hooks. A wrapper extends the current learner with added
functionality and expands the hyperparameter set of the learner with additional
hyperparameters provided by the wrapper. Currently, many wrappers are
available, e.g., missing value imputation, class imbalance correction,
feature selection, tuning, bagging and stacking, as well as a wrapper for
user-defined data pre-processing. Wrappers can be nested in other
wrappers, which can be used to create even more complex workflows.
The package also supports parallelization on different levels
\new{based on different parallelization backends (local multicore, socket, MPI)
with the package \code{parallelMap}~\citep{parallelMap} or on managed high-performance systems via the package \code{batchtools}~\citep{batchtools}.
Furthermore, visualization methods for research and teaching are also supplied.}

The \code{OpenML} package makes use of \code{mlr} as a supporting package. It
offers methods to automatically run \code{mlr} learners (flows) on OpenML
tasks while hiding all of the necessary structural transformations (see Section~\ref{sec:runs}).

\section{The OpenML R Package}
\label{sec:package}

The \code{OpenML} \code{R} package \citet{casalicchio17} is an interface to interact with
the OpenML server directly from within \code{R}. Users can retrieve
data sets, tasks, flows and runs from the server and also create and upload
their own. This section details how to install and configure the package
and demonstrates its most important functionalities.

\subsection{Installation and Configuration}
\label{sec:config}
To interact with the OpenML server, users need to authenticate using an \emph{API key},
a secret string of characters that uniquely identifies the user. It is generated and shown
on users' profile page after they register on the website \code{\url{http://www.openml.org}}. For demonstration
purposes, we will use a public read-only API key that only allows to retrieve
information from the server and should be replaced with the user's personal API key to be able to use all features.
The \code{R} package can be easily installed and configured as follows:

\begin{knitrout}\small
\definecolor{shadecolor}{rgb}{0.969, 0.969, 0.969}\color{fgcolor}\begin{kframe}
\begin{verbatim}
install.packages("OpenML")
library("OpenML")
saveOMLConfig(apikey = "c1994bdb7ecb3c6f3c8f3b35f4b47f1f")
\end{verbatim}
\end{kframe}
\end{knitrout}

The \code{saveOMLConfig} function creates a \code{config} file,
which is always located in a folder called \code{.openml} within the user's home
directory. This file stores the user's API key and other configuration settings, which
can always be changed manually or through the \code{saveOMLConfig} function.
Alternatively, the \code{setOMLConfig} function allows to set the API key and
the other entries \emph{temporarily}, i.e., only for the current \code{R} session.

\subsection{Listing Information}

In this section, we show how to list the available OpenML data sets, tasks, flows
and runs using listing functions that 
always return a \code{data.frame} containing the queried information.
Each data set, task, flow and run has a unique ID, which can be used to access it directly.

\paragraph{Listing Data Sets and Tasks:}
A list of all data sets and tasks that are available on the OpenML server
can be obtained using the \code{listOMLDataSets} and \code{listOMLTasks} function,
respectively.
Each entry provides information such as the ID, the name and basic
characteristics (e.g.,\ number of features, number of observations, classes, missing values) of the corresponding data set.
In addition, the list of tasks contains information about the task type (e.g., \code{"Supervised Classification"}), the evaluation measure
(e.g., \code{"Predictive Accuracy"}) and the estimation procedure (e.g., \code{"10-fold Crossvalidation"}) used to estimate model performance.
Note that multiple tasks can be defined for a specific data set, for example, the same data set can be used for multiple task types
(e.g. classification and regression tasks) as well as for tasks differing in their estimation procedure,
evaluation measure or target value. 

To find data sets or tasks that meet specific
requirements, one can supply arguments to the listing functions.
In the example below, we list all supervised classification tasks based on data sets having two classes for the target feature,
between 500 and 999 instances, at most 100 features and no missing values:

\begin{knitrout}\small
\definecolor{shadecolor}{rgb}{0.969, 0.969, 0.969}\color{fgcolor}\begin{kframe}
\begin{verbatim}
tasks = listOMLTasks(task.type = "Supervised Classification",
  number.of.classes = 2, number.of.instances = c(500, 999),
  number.of.features = c(1, 100), number.of.missing.values = 0)
tasks[1:2, c("task.id", "name", "number.of.instances", "number.of.features")]
##   task.id        name number.of.instances number.of.features
## 1      37    diabetes                 768                  9
## 2      49 tic-tac-toe                 958                 10
\end{verbatim}
\end{kframe}
\end{knitrout}

\paragraph{Listing Flows and Runs:}

When using the \code{mlr} package, flows are basically learners from 
\code{mlr}, which, as stated previously, 
can also be a more complex workflow when different \code{mlr} wrappers are nested.
Custom flows can be created by integrating custom machine learning algorithms and wrappers 
into \code{mlr}. 
The list of all available flows on OpenML can be downloaded using the \code{listOMLFlows} function.
Each entry contains information such as its ID, its name, its version and the user who
first uploaded the flow to the server.
Note that the list of flows will not only contain flows created with \code{R},
but also flows from other machine learning toolkits, such as WEKA \citep{Hall:2009p14495},
MOA~\citep{Bifet:2010p28524} and scikit-learn~\citep{scikit-learn}, 
which can be recognized by the name of the flow.

When a flow, along with a specific setup (e.g., specific hyperparameter values),
is applied to a task, it creates a run. The \code{listOMLRuns} function lists all runs
that, for example, refer to a specific \code{task.id} or \code{flow.id}. 
To list these evaluations as well, the \code{listOMLRunEvaluations} function can be used.
In Figure~\ref{fig:evalplot}, we used \code{ggplot2} \citep{ggplot2} to visualize 
the predictive accuracy of runs, for which only flows created with \code{mlr} 
were applied to the task with ID \code{37}:

\begin{knitrout}\small
\definecolor{shadecolor}{rgb}{0.969, 0.969, 0.969}\color{fgcolor}\begin{kframe}
\begin{verbatim}
res = listOMLRunEvaluations(task.id = 37, tag = "openml_r_paper")
res$flow.name = reorder(res$flow.name, res$predictive.accuracy)

library("ggplot2")
ggplot(res, aes(x = predictive.accuracy, y = flow.name)) + 
  geom_point() + xlab("Predictive Accuracy") + ylab("Flow Name")
\end{verbatim}
\end{kframe}\begin{figure}[!h]
\includegraphics[width=.95\textwidth]{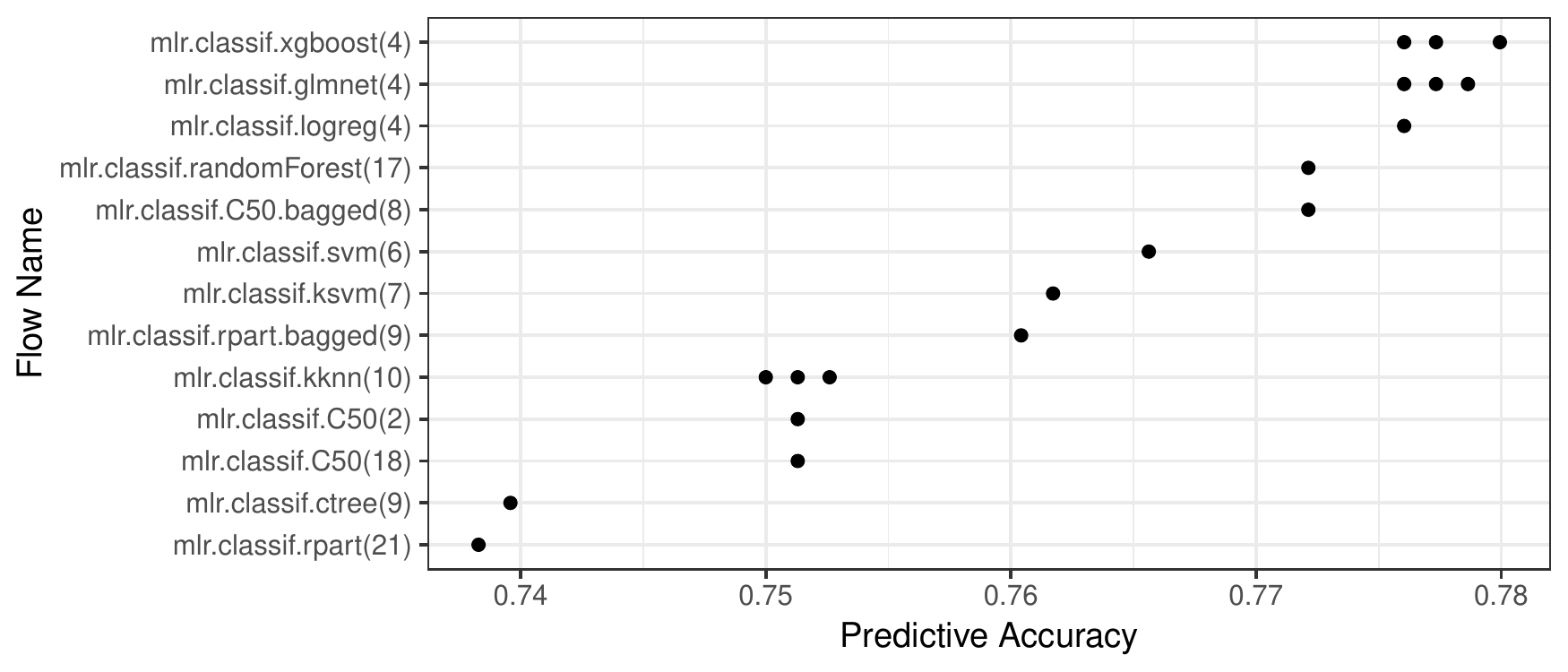} \caption[The predictive accuracy of some \code{mlr} flows on task 37]{The predictive accuracy of some \code{mlr} flows on task 37. The numbers in brackets refer to the version of the flow. Multiple dots for the same flow refer to runs with different hyperparameter values for that flow.}\label{fig:evalplot}
\end{figure}

\end{knitrout}


\subsection{Downloading OpenML Objects}

Most of the listing functions described in the previous section will list entities by
their OpenML IDs, e.g., the \code{task.id} for tasks, the \code{flow.id} for flows and the
\code{run.id} for runs. In this section, we show how these IDs can be used to
download a certain data set, task, flow or run from the OpenML server.
All downloaded data sets, tasks, flows and runs will be stored in the
\code{cachedir} directory, which will be in the \code{.openml} folder by default
but can also be specified in the configuration file (see Section~\ref{sec:config}).
Before downloading an OpenML object, the cache directory will be checked if that object
is already available in the cache. If so, no internet connection is necessary and the
requested object is retrieved from the cache.

\paragraph{Downloading Data Sets and Tasks:}
The \code{getOMLDataSet} function returns an \code{S3}-object of class
\code{OMLDataSet} that contains the data set as a \code{data.frame} in a
\code{\$data} slot, in addition to some pieces of meta-information:

\begin{knitrout}\small
\definecolor{shadecolor}{rgb}{0.969, 0.969, 0.969}\color{fgcolor}\begin{kframe}
\begin{verbatim}
ds = getOMLDataSet(data.id = 15)
ds
## 
## Data Set "breast-w" :: (Version = 1, OpenML ID = 15)
##   Default Target Attribute: Class
\end{verbatim}
\end{kframe}
\end{knitrout}

To retrieve tasks, the \code{getOMLTask} function can be used with their corresponding task ID.
Note that the ID of a downloaded task is not equal to the ID of the data set.
Each task is returned as an \code{S3}-object of class \code{OMLTask} and
contains the \code{OMLDataSet} object as well as the predefined
estimation procedure, evaluation measure and the target feature in an additional
\code{\$input} slot. Further technical information can be found in the package's
help page. 

\paragraph{Downloading Flows and Runs:}
The \code{getOMLFlow} function downloads all information of the flow, such as
the name, all necessary dependencies and all available hyperparameters that can be set.
If the flow was created in \code{R}, it can be converted into an \code{mlr} 
learner using the \code{convertOMLFlowToMlr} function:

\begin{knitrout}\small
\definecolor{shadecolor}{rgb}{0.969, 0.969, 0.969}\color{fgcolor}\begin{kframe}
\begin{verbatim}
mlr.lrn = convertOMLFlowToMlr(getOMLFlow(4782))
mlr.lrn
## Learner classif.randomForest from package randomForest
## Type: classif
## Name: Random Forest; Short name: rf
## Class: classif.randomForest
## Properties: twoclass,multiclass,numerics,factors,ordered,prob,class.weights
## Predict-Type: response
## Hyperparameters:
\end{verbatim}
\end{kframe}
\end{knitrout}

This allows users to apply the downloaded learner to other tasks or to modify 
the learner using functions from \code{mlr} and produce new runs.

The \code{getOMLRun} function downloads a single run and returns an \code{OMLRun} object
containing all information that are connected to this run, such as 
the ID of the task and the ID of the flow: 

\begin{knitrout}\small
\definecolor{shadecolor}{rgb}{0.969, 0.969, 0.969}\color{fgcolor}\begin{kframe}
\begin{verbatim}
run = getOMLRun(run.id = 1816245)
run
## 
## OpenML Run 1816245 :: (Task ID = 42, Flow ID = 4782)
## 	User ID  : 348
## 	Tags     : study_30
## 	Learner  : mlr.classif.randomForest(17)
## 	Task type: Supervised Classification
\end{verbatim}
\end{kframe}
\end{knitrout}

The most important information for reproducibility, next to the exact data set and flow version,
are the hyperparameter and seed settings that were used to create this run.
This information is contained in the \code{OMLRun} object and can be extracted via
\code{getOMLRunParList(run)} and \code{getOMLSeedParList(run)}, respectively.

If the run solves a supervised regression or classification 
task, the corresponding predictions can be accessed via \code{run\$predictions} and
the evaluation measures computed by the server via \code{run\$output.data\$evaluations}. 

\subsection{Creating Runs}
\label{sec:runs}
The easiest way to create a run is to define a learner, optionally with a preset 
hyperparameter value, using the \code{mlr} package. Each
\code{mlr} learner can then be applied to a specific \code{OMLTask} object
using the function \code{runTaskMlr}. This will create an \code{OMLMlrRun}
object, for which the results can be uploaded to the OpenML server as described
in the next section.
For example, a random forest from the \code{randomForest}
\code{R} package~\citep{randomForest} can be instantiated using the \code{makeLearner} 
function from \code{mlr} and can be applied to a classification task via:

\begin{knitrout}\small
\definecolor{shadecolor}{rgb}{0.969, 0.969, 0.969}\color{fgcolor}\begin{kframe}
\begin{verbatim}
lrn = makeLearner("classif.randomForest", mtry = 2)
task = getOMLTask(task.id = 37)
run.mlr = runTaskMlr(task, lrn)
\end{verbatim}
\end{kframe}
\end{knitrout}


To run a previously downloaded OpenML flow, one can use the 
\code{runTaskFlow} function, optionally with a list of hyperparameters:
\begin{knitrout}\small
\definecolor{shadecolor}{rgb}{0.969, 0.969, 0.969}\color{fgcolor}\begin{kframe}
\begin{verbatim}
flow = getOMLFlow(4782)
run.flow = runTaskFlow(task, flow, par.list = list(mtry = 2))
\end{verbatim}
\end{kframe}
\end{knitrout}

To display benchmarking results, one can use the \code{convertOMLMlrRunToBMR} function to convert
one or more \code{OMLMlrRun} objects to a single \code{BenchmarkResult} object from the
\code{mlr} package so that several powerful plotting functions 
\new{
(see \resizebox{\textwidth}{!}{\mbox{\url{http://mlr-org.github.io/mlr-tutorial/release/html/benchmark_experiments}}} for examples) 
}
from \code{mlr} can be applied to that object (see, e.g., Figure \ref{fig:bmrplot}).

\subsection{Uploading and Tagging}

\paragraph{Uploading OpenML Objects:}
It is also possible to upload data sets, flows and runs
to the OpenML server to share and organize experiments and results
online. Data sets, for example, are uploaded with the \code{uploadOMLDataSet} function.
OpenML will \emph{activate} the data set if it passes all checks, meaning that it will be
returned in listing calls.
Creating tasks from data sets is currently only possible through
the website, see \url{http://www.openml.org/new/task}.

\code{OMLFlow} objects can be uploaded to the server with the \code{uploadOMLFlow} 
function and are automatically versioned by the server: when a learner is uploaded carrying
a different \code{R} or package version, a new version number and \code{flow.id} is assigned.
If the same flow has already been uploaded to the server, a message that the flow already exists is
displayed and the associated \code{flow.id} is returned. Otherwise, the flow is uploaded 
and a new \code{flow.id} is assigned to it:

\begin{knitrout}\small
\definecolor{shadecolor}{rgb}{0.969, 0.969, 0.969}\color{fgcolor}\begin{kframe}
\begin{verbatim}
lrn = makeLearner("classif.randomForest")
flow.id = uploadOMLFlow(lrn)
\end{verbatim}
\end{kframe}
\end{knitrout}

A run created with the \code{runTaskMlr} or the \code{runTaskFlow} function can
be uploaded to the OpenML server using the \code{uploadOMLRun} function. The server
will then automatically compute several evaluation measures for this run, which
can be retrieved using the \code{listOMLRunEvaluations} function as described previously.

\paragraph{Tagging and Untagging OpenML Objects:}

The \code{tagOMLObject} function is able to tag data sets, tasks, flows and
runs with a user-defined string, so that finding OpenML objects with a
specific tag becomes easier. For example, the task with ID 1 can be tagged as follows:

\begin{knitrout}\small
\definecolor{shadecolor}{rgb}{0.969, 0.969, 0.969}\color{fgcolor}\begin{kframe}
\begin{verbatim}
tagOMLObject(id = 1, object = "task", tags = "test-tagging")
\end{verbatim}
\end{kframe}
\end{knitrout}

To retrieve a list of objects with a given tag, the
\code{tag} argument of the listing functions can be used 
(e.g., \code{listOMLTasks(tag = "test-tagging")}). 
The listing functions for data sets, tasks, flows and runs also show the tags 
that were already assigned, for example, we already tagged data sets from 
UCI~\citep{Asuncion:2007p519} with the string \code{"uci"} so that they can be 
queried using \code{listOMLDataSets(tag = "uci")}.
In order to remove one or more tags from an \code{OpenML} object, the
\code{untagOMLObject} function can be used, however, only self-created tags can be removed, e.g.:

\begin{knitrout}\small
\definecolor{shadecolor}{rgb}{0.969, 0.969, 0.969}\color{fgcolor}\begin{kframe}
\begin{verbatim}
untagOMLObject(id = 1, object = "task", tags = "test-tagging")
\end{verbatim}
\end{kframe}
\end{knitrout}

\subsection{Further Features}
Besides the aforementioned functionalities, the \code{OpenML} package allows
to fill up the cache directory by downloading multiple objects at once (using the
\code{populateOMLCache} function), to remove all files from the cache directory 
(using \code{clearOMLCache}), to get the current status 
of cached data sets (using \code{getCachedOMLDataSetStatus}), to delete OpenML 
objects created by the uploader (using \code{deleteOMLObject}), to list all 
estimation procedures (using \code{listOMLEstimationProcedures}) as well as all 
available evaluation measures (using \code{listOMLEvaluationMeasures}) and to get 
more detailed information on data sets (using \code{getOMLDataSetQualities}). 

\section{Case Study}
\label{sec:study}

In this section, we illustrate the usage of OpenML by performing a small
comparison study between a random forest, bagged trees and single classification trees. We first
create the respective binary classification learners using \code{mlr}, then query
OpenML for suitable tasks, apply the learners to the tasks and finally evaluate
the results.

\subsection{Creating Learners}

We choose three implementations of different tree
algorithms, namely the \textit{CART} algorithm implemented in the
\code{rpart} package~\citep{rpart}, the \textit{C5.0} algorithm from the package
\code{C50}~\citep{C50} and the \textit{conditional inference trees} implemented
in the \code{ctree} function from the package \code{party}~\citep{party}.
For the \textit{random forest}, we use the implementation from the package
\code{randomForest}~\citep{randomForest}.
The bagged trees can conveniently be created using \code{mlr}'s bagging wrapper.
\new{Note that we do not use bagging for the \code{ctree} algorithm due to large memory requirements.}
For the random forest and all bagged tree learners, the number of trees is set to 50.
We create a list that contains the random forest, the two bagged trees and the three tree algorithms:

\begin{knitrout}\small
\definecolor{shadecolor}{rgb}{0.969, 0.969, 0.969}\color{fgcolor}\begin{kframe}
\begin{verbatim}
lrn.list = list(
  makeLearner("classif.randomForest", ntree = 50),
  makeBaggingWrapper(makeLearner("classif.rpart"), bw.iters = 50),
  makeBaggingWrapper(makeLearner("classif.C50"), bw.iters = 50),
  makeLearner("classif.rpart"),
  makeLearner("classif.C50"),
  makeLearner("classif.ctree")
)
\end{verbatim}
\end{kframe}
\end{knitrout}

\subsection{Querying OpenML}

For this study, we consider only binary classification tasks that use smaller 
data sets from UCI~\citep{Asuncion:2007p519}, e.g., between 100 and 999 observations, 
have no missing values and use 10-fold cross-validation for validation:

\begin{knitrout}\small
\definecolor{shadecolor}{rgb}{0.969, 0.969, 0.969}\color{fgcolor}\begin{kframe}
\begin{verbatim}
tasks = listOMLTasks(data.tag = "uci",
  task.type = "Supervised Classification", number.of.classes = 2,
  number.of.missing.values = 0, number.of.instances = c(100, 999),
  estimation.procedure = "10-fold Crossvalidation")
\end{verbatim}
\end{kframe}
\end{knitrout}

Table \ref{tab:tasks} shows the resulting tasks of the query, which will be used for the further analysis.

\begin{table}[ht]
\centering
\begin{tabular}{cccc}
  \hline
task.id & name & number.of.instances & number.of.features \\ 
  \hline
 37 & diabetes & 768 &   9 \\ 
   39 & sonar & 208 &  61 \\ 
   42 & haberman & 306 &   4 \\ 
   49 & tic-tac-toe & 958 &  10 \\ 
   52 & heart-statlog & 270 &  14 \\ 
   57 & ionosphere & 351 &  35 \\ 
   \hline
\end{tabular}
\caption{Overview of OpenML tasks that will be used in the study.} 
\label{tab:tasks}
\end{table}

\subsection{Evaluating Results}
\label{sec:evaluation}
We now apply all learners from \code{lrn.list} to the selected tasks using the \code{runTaskMlr} function
and use the \code{convertOMLMlrRunToBMR} function to create a single 
\code{BenchmarkResult} object containing the results of all experiments.
This allows using, for example, the \code{plotBMRBoxplots} function from 
\code{mlr} to visualize the experiment results (see Figure~\ref{fig:bmrplot}):

\begin{knitrout}\small
\definecolor{shadecolor}{rgb}{0.969, 0.969, 0.969}\color{fgcolor}\begin{kframe}
\begin{verbatim}
grid = expand.grid(task.id = tasks$task.id, lrn.ind = seq_along(lrn.list))
runs = lapply(seq_row(grid), function(i) {
  task = getOMLTask(grid$task.id[i])
  ind = grid$lrn.ind[i]
  runTaskMlr(task, lrn.list[[ind]])
})
bmr = do.call(convertOMLMlrRunToBMR, runs)
plotBMRBoxplots(bmr, pretty.names = FALSE)
\end{verbatim}
\end{kframe}
\end{knitrout}

\begin{knitrout}\small
\definecolor{shadecolor}{rgb}{0.969, 0.969, 0.969}\color{fgcolor}\begin{figure}[!h]

{\centering \includegraphics[width=\textwidth]{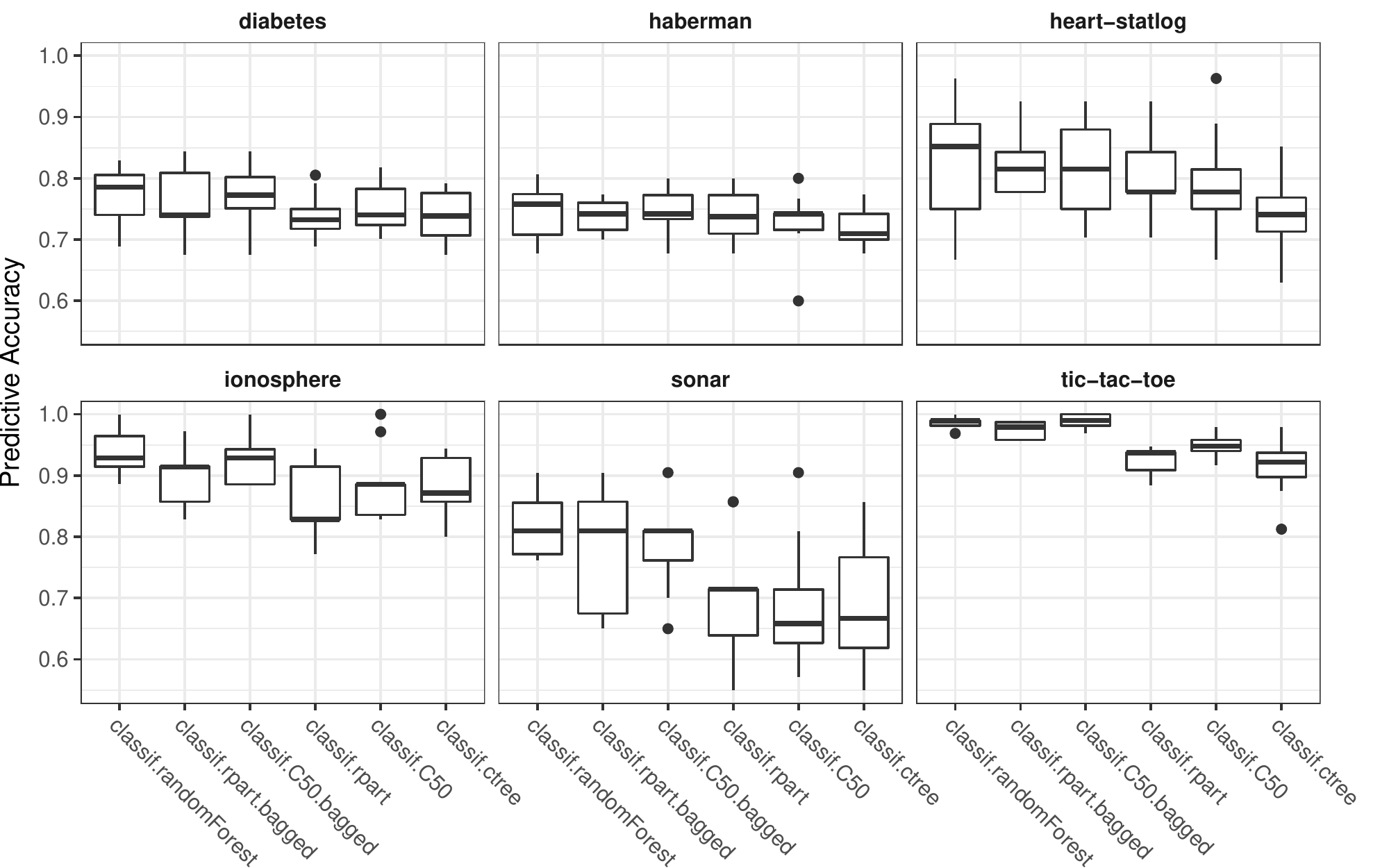} 

}

\caption[Cross-validated predictive accuracy per learner and task]{Cross-validated predictive accuracy per learner and task. Each boxplot contains 10 values for one complete cross-validation.}\label{fig:bmrplot}
\end{figure}

\end{knitrout}

We can upload and tag the runs, e.g., with the string \code{"study\_30"}
to facilitate finding and listing the results of the runs using this tag:

\begin{knitrout}\small
\definecolor{shadecolor}{rgb}{0.969, 0.969, 0.969}\color{fgcolor}\begin{kframe}
\begin{verbatim}
lapply(runs, uploadOMLRun, tags = "study_30")
\end{verbatim}
\end{kframe}
\end{knitrout}

The server will then compute all possible measures, which takes some time
depending on the number of runs.
The results can then be listed using the \code{listOMLRunEvaluations} function
and can be visualized using the \code{ggplot2} package: 

\begin{knitrout}\small
\definecolor{shadecolor}{rgb}{0.969, 0.969, 0.969}\color{fgcolor}\begin{kframe}
\begin{verbatim}
evals = listOMLRunEvaluations(tag = "study_30")
evals$learner.name = as.factor(evals$learner.name)
evals$task.id = as.factor(evals$task.id)

library("ggplot2")
ggplot(evals, aes(x = data.name, y = predictive.accuracy, colour = learner.name, 
  group = learner.name, linetype = learner.name, shape = learner.name)) +
  geom_point() + geom_line() + ylab("Predictive Accuracy") + xlab("Data Set") +
  theme(axis.text.x = element_text(angle = -45, hjust = 0))
\end{verbatim}
\end{kframe}\begin{figure}[!h]

{\centering \includegraphics[width=\textwidth]{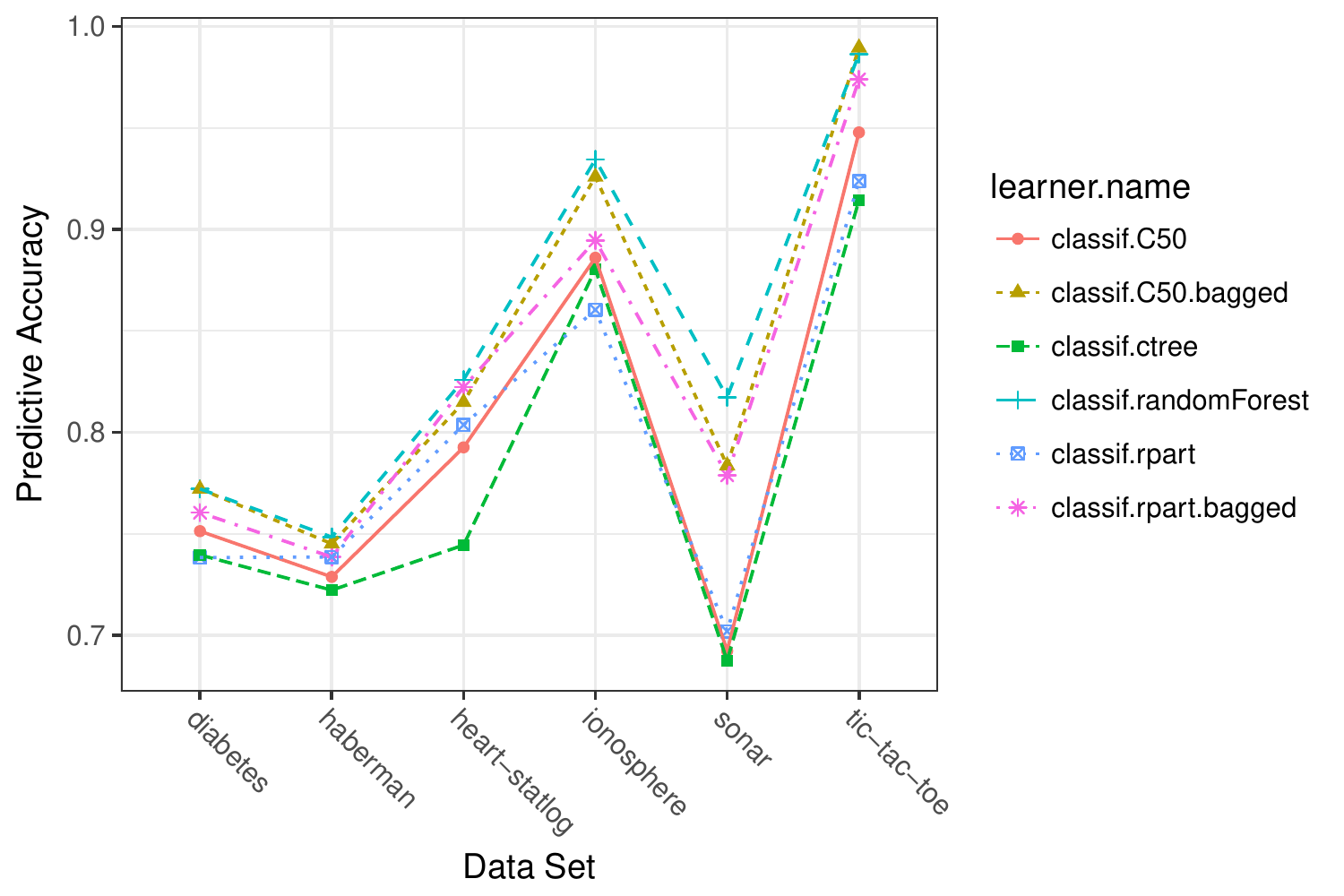} 

}

\caption[Results of the produced runs]{Results of the produced runs. Each point represents the averaged predictive accuracy over all cross-validation iterations generated by running a particular learner on the respective task.}\label{fig:res}
\end{figure}

\end{knitrout}

Figure~\ref{fig:res} shows the cross-validated predictive accuracies of our six learners on the considered tasks. Here, the random forest produced the best predictions, except on \new{the tic-tac-toe data set}, where the bagged C50 trees achieved a slightly better result. In general, the two bagged trees performed marginally worse than the random forest and better than the single tree learners.

\section{Conclusion and Outlook}
\label{sec:outlook}

OpenML is an online platform for open machine learning that is aimed at
connecting researchers who deal with any part of the machine learning
workflow. The OpenML platform automates the sharing of machine learning tasks and 
experiments through the tools that scientists are already using, such as \code{R}. 
The \code{OpenML} package introduced in this paper makes it easy to share and
reuse data sets, tasks, flows and runs directly from the current \code{R} session 
without the need of using other programming environments or the web interface.

Current work is being done on implementing the possibility to connect to OpenML 
via browser notebooks \new{(\url{https://github.com/everware})} and running analysis 
directly on online servers without the need of having \code{R} or any other 
software installed locally. In the future, it will also be possible that users 
can specify with whom they want to share, e.g., data sets.

\bibliographystyle{spbasic}      
\bibliography{Bib}

\end{document}